\RequirePackage{fix-cm}
\documentclass[smallextended]{svjour3}       
\smartqed  
\usepackage{amsmath,amsfonts}
\usepackage{algorithmic}
\usepackage{algorithm}
\usepackage[caption=false,font=normalsize,labelfont=sf,textfont=sf]{subfig}
\usepackage{stfloats}
\usepackage{url}
\usepackage{graphicx}
\usepackage{cite}
\usepackage{url}
\usepackage{xcolor}
\begin{document}

\title{Robust and Fast Vehicle Detection using Augmented Confidence Map 
}

\author{Hamam Mokayed \and Palaiahnakote Shivakumara \and
        Lama Alkhaled \and Rajkumar Saini \and Muhammad Zeshan Afzal
        \and Yan Chai Hum \and Marcus Liwicki
}


\institute{Hamam Mokayed \at
              \email{Hamam.Mokayed@ltu.se}           
           \and
           Palaiahnakote Shivakumara \at
              \email{shiva@um.edu.my} 
            \and
           Lama Alkhaled \at
              \email{lama.alkhaled@ltu.se} 
              \and
           Rajkumar Saini \at
              \email{rajkumar.saini@ltu.se} 
              \and
            Muhammad Zeshan Afzal \at
              \email{muhammad\_zeshan.afzal@dfki.de}
               \and
            Yan Chai Hum \at
              \email{humyc@utar.edu.my}
              \and
            Marcus Liwicki \at
              \email{marcus.liwicki@ltu.se}
}

\date{Received: date / Accepted: date}

\maketitle

\begin{abstract}
Vehicle detection in real-time scenarios is challenging because of the time constraints and the presence of multiple types of vehicles with different speeds, shapes, structures, etc. This paper presents a new method relied on generating confidence map-for robust and faster vehicle detection. To reduce the adverse effect of different speeds, shapes, structures, and the presence of several vehicles in a single image, we introduce the concept of augmentation which highlights the region of interest containing the vehicles. The augmented map is generated by exploring the combination of multiresolution analysis and maximally stable extremal regions (MR-MSER). The output of MR-MSER is supplied to fast CNN to generate a confidence map, which results in candidate regions. Furthermore, unlike existing models that implement complicated models for vehicle detection, we explore the combination of a rough set and fuzzy-based models for robust vehicle detection. To show the effectiveness of the proposed method, we conduct experiments on our dataset captured by drones and on several vehicle detection benchmark datasets, namely, KITTI and UA-DETRAC. The results on our dataset and the benchmark datasets show that the proposed method outperforms the existing methods in terms of time efficiency and achieves a good detection rate.
\end{abstract}

\section{Introduction}
\label{intro}
With the increasing importance of vehicle-related industrial applications, vehicle detection has been a very interesting research topic for the last two decades \cite{Ref1}. The research started based on applying traditional nondeep learning methods to accomplish the task. Because the implemented features of these conventional algorithms are susceptible to image deformation, vehicle occlusion, and illumination changes, these methods have lower accuracy and can be deployed in limited scenarios \cite{Ref2,Ref4,Ref8}. With the revolution of deep learning, many methods have been suggested to enhance vehicle detection performance by proposing complex network architectures \cite{Ref5,Ref6,Ref7}. The main drawback of these methods is the cost of the high computational power machines and longer processing time requirements. Considerable research related to object detection, in general, has been conducted, and many of these generic classifiers cannot achieve competitive performance on vehicle detection benchmarks because of the different problems related to vehicle detection, such as large variations in light, dense occlusion, and size differences. These factors motivated us to propose a new method that combines a fast deep learning method for augmented feature maps with traditional methods for vehicle detection. Fig. 1 illustrates the contribution of the proposed model to detect multiple vehicles in drone images. The input drone image containing multiple vehicles is shown in Fig. 1(a). The region of interest is detected by the combination of multiresolution and maximally stable extremal regions (MR-MSE) as shown in Fig. 1(b). The regions are subjected to a CNN to generate an augmented confidence map (ACM), which outputs saliency for the vehicle regions seen in Fig. 1(c). The combination of rough set and fuzzy set is proposed to classify nonvehicle and vehicle blobs, as shown in Fig. 1(d)-(e).
\begin{figure}
  \includegraphics{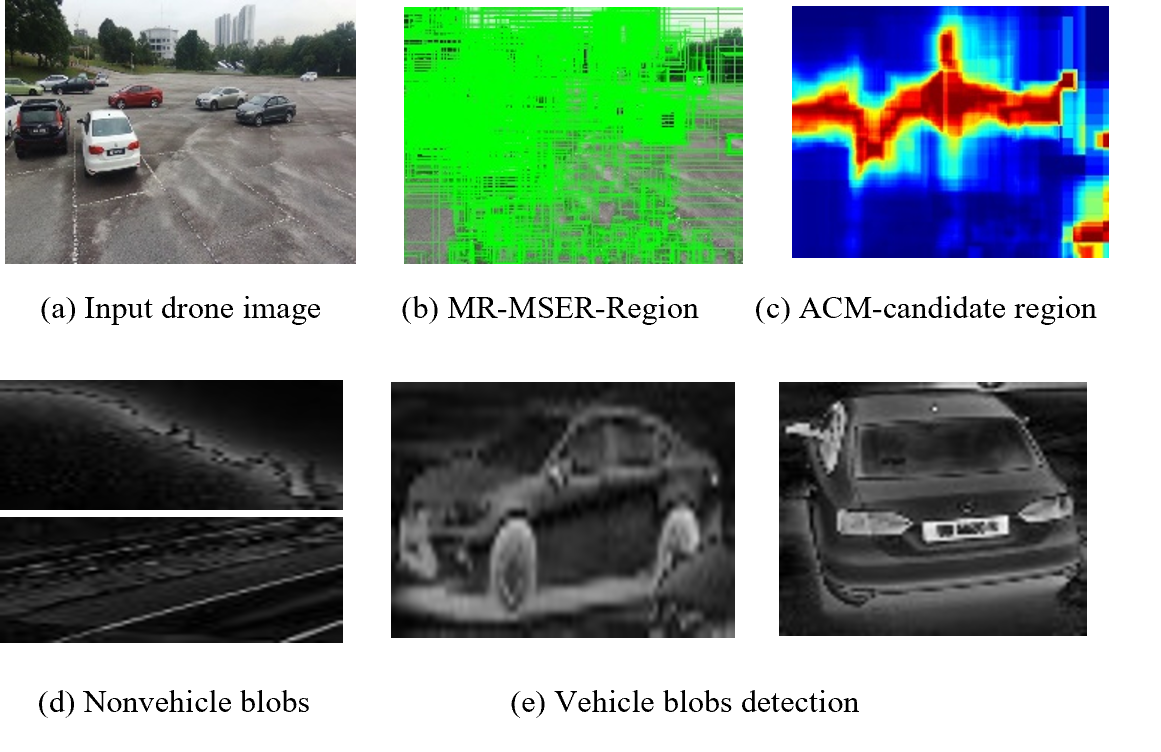}
\caption{The proposed steps for vehicle detection in multivehicle environment}
\label{fig:1}       
\end{figure}

\section{Related Work}
\label{sec:1}
Several methods have been developed in the past for addressing the challenges of vehicle detection, especially for tiny vehicles and multiple vehicles in images, based on a deep learning approach. Wang et al. \cite{Ref9} presented a review of vehicle detection for intelligent vehicles. This paper discusses various approaches for addressing different challenges of vehicle detection. It is noted from the review that robust vehicle detection in crowded scenes and when the image contains multiple vehicles are still considered open challenges for vehicle detection. For example, He et al. \cite{Ref10} proposed a method for vehicle detection with deep features and shape features. The approach uses the combination of R-CNN and SVM for vehicle detection in satellite images. Hu et al. \cite{Ref7} used a scale-intensive CNN for fast vehicle detection. The technique targets small-scale vehicles and uses contextual information for vehicle detection. Other researchers proposed to develop a YOLO-based method for front-end vehicle detection \cite{Ref11}. The aim of the method is to achieve time efficiency to meet the requirements of real-time applications. Li et al. \cite{Ref13} explored the combination of several deep learning models for detecting tiny vehicles in images. Since the target of the method is to reduce the number of computations, the approach uses the YOLOv3 architecture for vehicle detection. Ojha et al.\cite{Ref14} proposed an instance-based segmentation-based model for vehicle detection. To achieve this, the method uses R-CNN transfer learning. Shan et al. \cite{Ref15} explored instance segmentation for real-time vehicle detection. The method estimates the precise loss for predicting the bounding box regression for vehicle detection. Shi et al. \cite{Ref16} introduced orientation-aware vehicle detection based on an anchor-free object detection approach for aerial images. The method splits transfer learning into multitask learning for vehicle detection. Wang et al. \cite{Ref17} proposed a model for vehicle detection based on single- and two-stage deep learning techniques. The method uses a soft weighted average method to achieve the best results. Yao et al. \cite{Ref18} proposed YOLOv3 for traffic vehicle detection. The approach uses a combination of clustering and deep learning to overcome the challenges of traffic vehicle detection.
It is observed from a review of different vehicle detection methods that most of the methods focus on high-resolution images and images captured in a head-on direction but not drone images. Therefore, one can infer that the existing method may not be effective and robust for drone images because of the adverse effects of height distance variation, oblique angles, and weather conditions. Thus, this study aims to develop a robust approach for accurately detecting vehicles in drone images as well as normal images by exploring the combination of deep learning, augmented confidence maps, rough sets and fuzzy logic. Hence, the contributions are threefold: (i) generating augmented confidence maps with the help of multiresolution and maximally stable extremal regions, (ii) exploring the combination of rough set and fuzzy for vehicle blobs classification, and (iii) the way the proposed work integrates the deep learning, augmentation, rough set and fuzzy is new compared to the state-of-the-art methods.
\section{Proposed Method}
\label{sec:2}
This research aims to propose a robust vehicle detection method by combining a fast deep learning method with traditional steps. The work can be divided into two main stages, namely, augmented confidence map (ACM) and a traditional classifier based on ACM.
\subsection{Augmented confidence map (ACM)}
\label{sec:3}
First, a robust algorithm, namely, maximally stable extremal regions (MSER), has been proposed to provide a new set of image elements. These elements are known as extremal regions, with two features extracted from the projective transformation of image coordinates and the monotonic transformation of image intensities \cite{Ref19}. Affine invariant feature descriptors are computed on a grayscale image, and the robustness of MSER is different from multiple measurement regions picked up from invariant constructions from extremal regions \cite{Ref20}, but some of the regions show discriminative properties that are significantly larger and possibly used to establish tentative correspondences \cite{Ref19}. Since MSER can generate many blobs with different sizes based on detection in an original image resolution, the different resolutions that can yield from long distances or blurred (coarse image), the loss of details in the image and different regions are connected to their neighbor \cite{Ref20}. The MSER can work better in invariance to changing scale in the scene image at different resolutions, which achieves stable vehicle regions. The resolution of an image is obtained by a scale pyramid (without a Gaussian filter) with one octave for a scale and a total of three scales from the finest image (input image) to the coarser image (blurred image). This will lead to generating multiresolution maximally stable extremal regions MR-MSER, and this MR-MSER is applied to vehicle images with 640×480 resolution \cite{Ref19}.
\begin{figure}
  \includegraphics{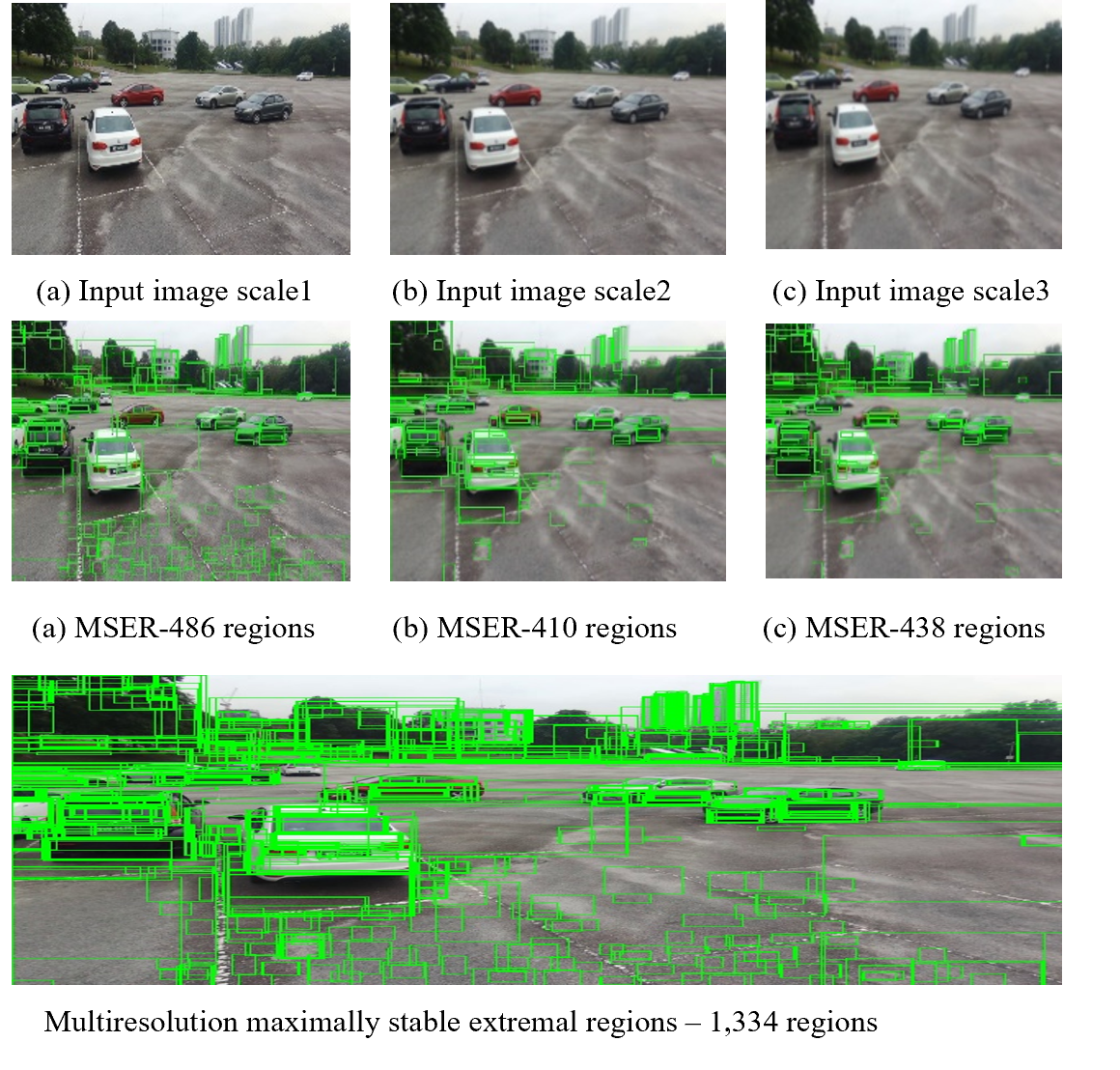}
\caption{The regions generated by applying MSER and MR-MSER over vehicle images}
\label{fig:2}       
\end{figure}
An augmentation process is applied to the MR-MSER to enhance the detection recall of the regions. The augmentation pipeline is adopted in each MR-MSER region shown in Fig. 3. Each MR-MSER with a different size is in a rectangular bounding box without moving its center. The bounding box is then squarified by inflating 30\% and 60\% of its area in three different dimensions. The different squared dimensions are resized to a 28×28 pixel image patch. Each image patch applied online jitter is randomly rotated within $[-\pi/4,~\pi/4]$ for four different times, which will be used in model training \cite{Ref21}.
\begin{figure}
  \includegraphics{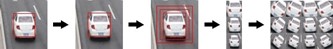}
\caption{The applied augmentation pipeline}
\label{fig:3}       
\end{figure}

The data generated by the augmented MR-MSER is used to generate the confidence map. The data are extracted from 100 images randomly selected from the KITTI dataset, UA-DETRAC dataset, and our collected dataset. The annotation of the ground truth (foreground, which is the vehicle) is performed on 100 images.
\begin{figure}
  \includegraphics{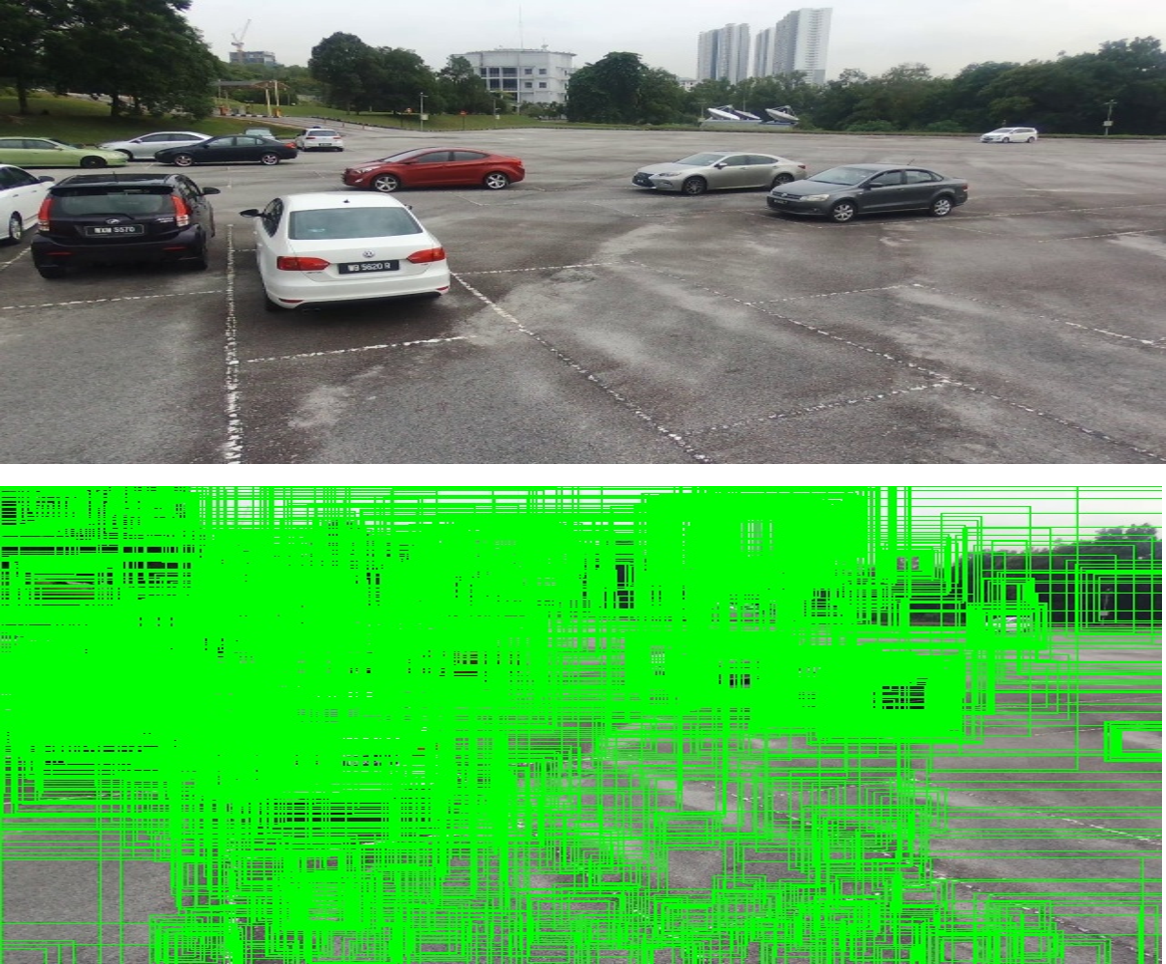}
\caption{The regions generated by applying augmented MR-MSER over vehicle images}
\label{fig:4}       
\end{figure}
By using the augmented MR-MSER, the positive samples are obtained by the annotated ground-truth vehicle and having an intersection-over-union IoU>0.5, no less than one of the ground-truth vehicle annotations. The negative samples are equal to positive samples that are randomly selected with IoU=0 for every ground-truth vehicle annotation. The proposed framework uses a slight variant of LeNet for classifying image patches from augmented MR-MSER in two classes, either foreground (vehicle) or background (nonvehicle). The input is a 28×28 pixel image. The framework used to generate the confidence map has a total of three convolutional layers with 128,256,512 feature maps and two fully connected layers that have 512 units. The kernel size and stride are fixed to 3 and 1, respectively, in each convolutional layer. Max pooling with a 2×2 pooling size is performed after each convolutional layer. The rectified linear unit (ReLU) is performed as an activation function for all layers except the prediction layer (last layer) using the sigmoid function. A dropout layer is added with 0.5 rates before the prediction layer to increase the model accuracy. The CNN model is trained with a training dataset that is extracted from 100 images based on the MR-MSER that has the threshold on intersection-over-union IoU such that the set for the foreground (vehicle) and the set for the background (nonvehicle) is obtained with equal size. The training dataset is split into a training set and validation set with an appropriate ratio, and the model is trained in minibatches with the optimum batch size and a number of epochs. The learning optimization in training the model uses the Adam optimizer, which is straightforward in implementation, efficient in computation with fewer memory requirements, invariant to diagonal rescaling of the gradients, and well suited for larger data and/or parameters \cite{Ref22}. Since the model is trained for binary classification, binary cross-entropy is the loss function for the model. The confidence map is computed by using the confidence values obtained from the stacking of each augmented MR-MSER proposal. The confidence map described as higher intensity regions likely denotes the interest of vehicle components. The threshold is used in the confidence map to filter the region that likely does not belong to the foreground regions. The remaining foreground regions are used for further image denoising and segmentation.
\begin{figure}
  \includegraphics{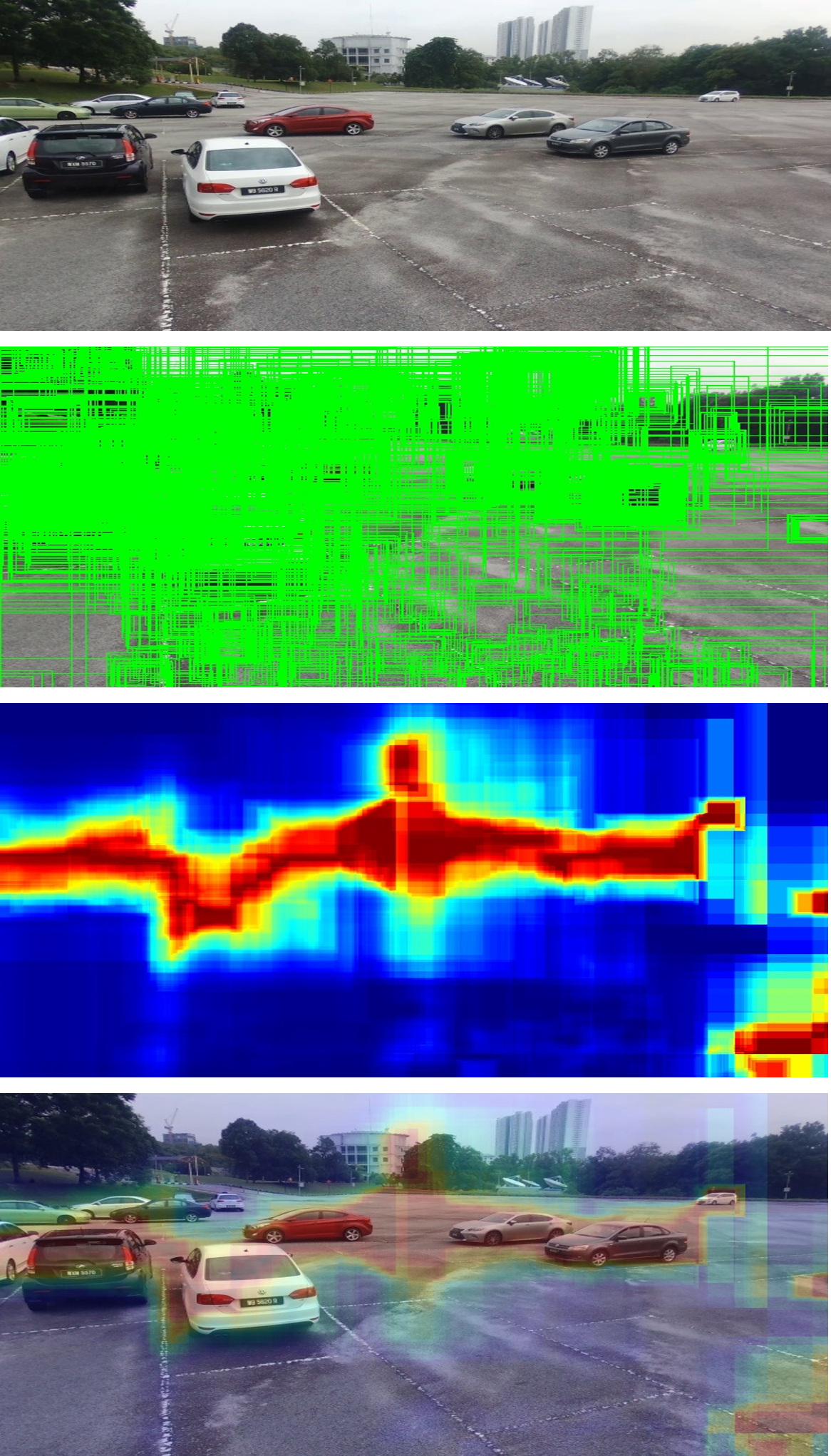}
\caption{Augmented confidence map}
\label{fig:5}       
\end{figure}

\subsection{Traditional classifier based on ACM}
\label{sec:4}
The purpose of using the rough set approach is knowledge discovery and approximation of sets using granular information. Given a confidence map, the process for granularization splits the input image window into multiple windows with a resolution of each subwindow (g = 4). The main purpose is to classify pixel values into vehicle and nonvehicle approximations. Let a set of objects be U. There is also an indiscernibility relation $R\subseteq~U*U$ that refers to the central concept of rough set theory. In the indiscernibility relation, the values of the object are identical considering a subset of the related attributes. In other words, it is an equivalence relation where all identical values of the object are elementary. Hence, R can also be considered an equivalence relation. Let X be a subset of U with two possibilities, either is crisp which is explicit with respect to R if the boundary region of X is empty, or is rough, which is inexplicit with respect to R if the boundary region of X is nonempty using RST to characterize the set U as possible for lower approximation and upper approximation, and boundary region of set X.

\begin{equation}
\underline{R}(X)= \cup_{x\in~U} R(X):R(X)\subseteq X
\end{equation}
R-upper approximation of X:
\begin{equation}
\overline{R}(X)= \cup_{x\in~U} R(X):R(X) \cap X \ne \phi
\end{equation}
R-boundary region of X:
\begin{equation}
RN_R(X)= \overline{R}(X)-\underline{R}(X)
\end{equation}
Rough entropy (RE) is introduced, which can avoid imprecision to find the optimum threshold as precisely as possible. The rough entropy threshold (RET) as the reference for the threshold in the binarization approach in the grayscale image, which was obtained by using a sliding window with a nonoverlapping granule window in m×n, is set as a 2×2 window size. RET can be defined as \cite{Ref23}
\begin{equation}
RE_T=-\frac{e}{2}[R_{OT}\log(R_{OT})+R_{BT}\log(R_{BT})]
\end{equation}
where\\
$R_{OT}=1- \mid \overline{O}_{T}\mid/\mid \underline{O}_{T}\mid$ is the roughness of the object \\ 
$R_{BT}=1- \mid \overline{B}_{T}\mid/\mid \underline{B}_{T}\mid$is the roughness of the background \\
$\mid \overline{O}_{T}\mid$ and $\mid \underline{O}_{T}\mid$ are the cardinality of the sets $\overline{O}_{T}$ and $\underline{O}_{T}$ for a given image depending on the value T.\\
$\mid \overline{B}_{T}\mid$ and $\mid \underline{B}_{T}\mid$ are the cardinality of the sets $\overline{B}_{T}$ and $\underline{B}_{T}$ for a given image depending on the value T. \\

The principle of reducing the roughness of both the object and background and maximizing $RE_T$ is computed for every T representing the object and background regions, respectively (0,…,T) and (T+1,…,L-1). The optimum threshold is selected for the maximum $RE_T$ to provide the object-background segmentation given by the definition of $T^*$
\begin{equation}
T^* = arg\:max_T RE_T
\end{equation}

Note that maximizing the rough entropy $RE_T$ to obtain the required threshold basically implies minimizing both the object roughness and background roughness such that this method is an object enhancement/extraction method \cite{Ref23}
\begin{figure}
  \includegraphics{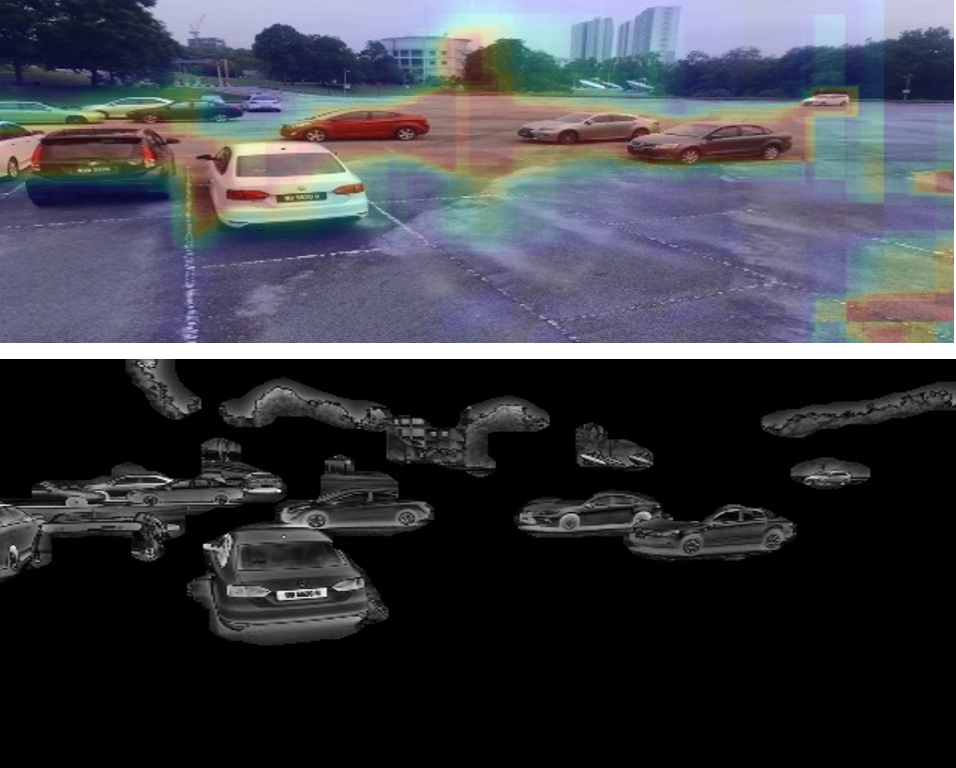}
\caption{Applying rough set theory to the ACM}
\label{fig:6}       
\end{figure}

Finally, the fuzzy classifier assigns a class label to an object based on the object description. This can be known as predicting the class label where the object description is in the form of a vector. This vector contains values of the features deemed to be relevant information for the classification process. The feature is the intensity distribution in the ACM.The classifier has learned with a training dataset to predict class labels whether it is vehicle or noise. A sample of the collected data used for training is shown in the figure below.

\begin{figure}
  \includegraphics{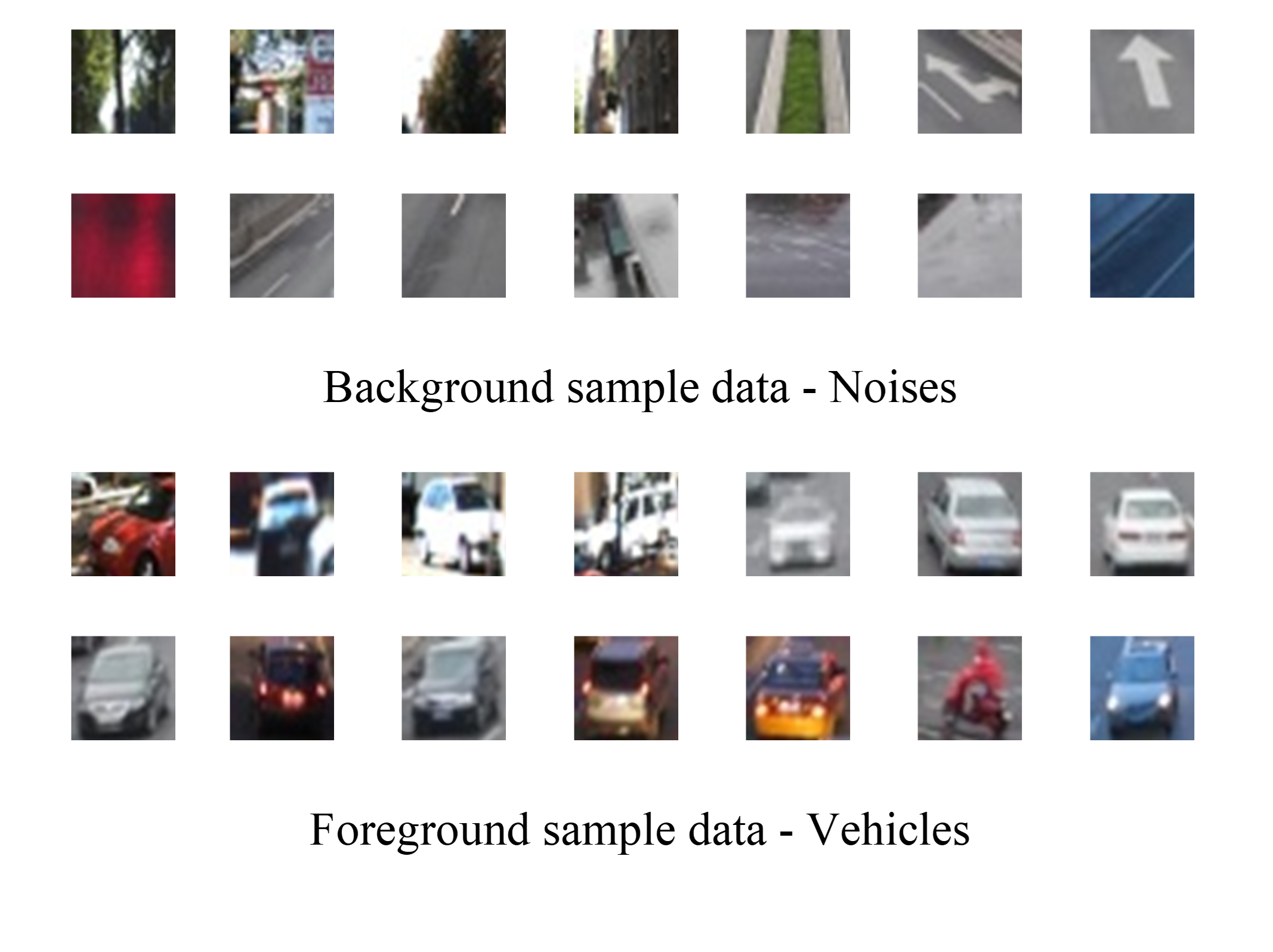}
\caption{Collected data used for the fuzzy logic training process}
\label{fig:7}       
\end{figure}

Fuzzy ruble-based classifiers have utilized soft labeling. A standard assumption in pattern recognition is that the classes are mutually exclusive. Soft labeling with the given dataset is split into 2 classes: vehicle set and background set. Both the vehicle set and background set are computed with the intensity distribution and soft labeling is performed for different sets. The representation of the output image will be in bicolor, the “green” color is labeled as a vehicle, and the “red” color is labeled as background, as shown in Fig. 8.

\begin{figure}
  \includegraphics{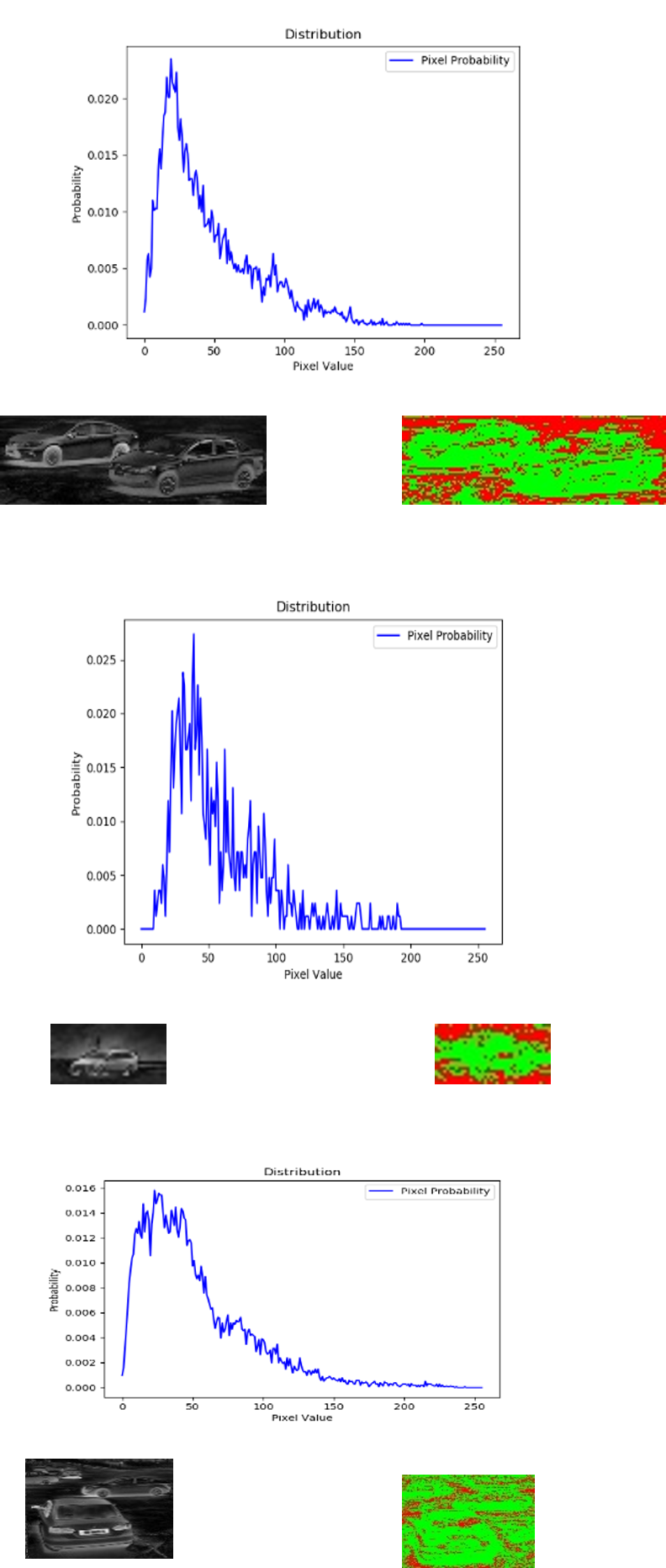}
\caption{Results of fuzzy logic over the nominated blobs}
\label{fig:8}       
\end{figure}

\section{Experimental Results}
\label{sec:5}
To evaluate the proposed system, first, we used two standard datasets mainly used for vehicle-related problems named KITTI and UA-DETRAC. In addition, a self-collected dataset captured from drones is used for further evaluation. The self-collected dataset consists of images collected at different times (early morning, afternoon, and evening) to obtain vehicle images with multiple densities from the open car parking area and highway. In addition, the dataset includes images captured from varying height distances from 1-3, 3-5, and more than 7 meters at different angles. The overall number of collected images is 1,000. A sample of our images is shown in the figure below.
\begin{figure}
  \includegraphics{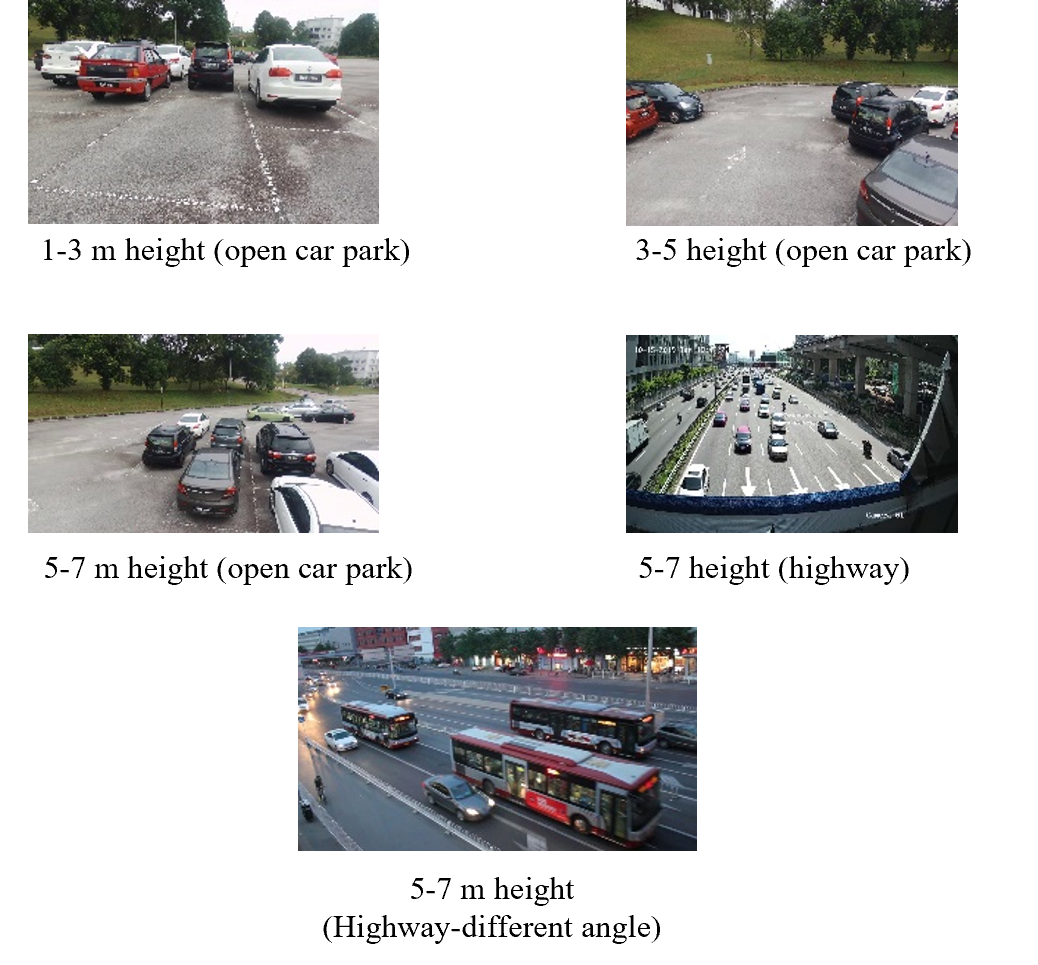}
\caption{Samples of our collected dataset for the vehicle detection problem}
\label{fig:9}       
\end{figure}
\subsection{Ablation study}
\label{sec:6}
In this work, the combination of an ACM and a traditional classifier is proposed to reach a promising level of accuracy and performance for diversified situations. To show the effectiveness of the proposed steps in the conducted work, the following experiments were implemented toward ablation studies on our collected data and the other two benchmarked data. The results reported in Table I show the importance of applying the augmentation process on MR-MSER to enhance the detection performance of the regions. The three experiments shown in Tables I, II, and III justify the need for ACM for better generalization ability, flexibility, and robustness to different situations compared to applying only MSER or MR-MSER at the ACM stage.
\begin{table}
\caption{EVALUATION OF VEHICLE DETECTION USING THE KITTI DATASET}
\label{tab:1}       
\begin{tabular}{lll}
\hline\noalign{\smallskip}
Measures & AP & Input image size  \\
\noalign{\smallskip}\hline\noalign{\smallskip}
Method with MSER & 97.22 & 512*512 \\
Method with MR-MSER & 84.45 & 512*512 \\
ACM proposed work & 90.08 & 512*512 \\
\noalign{\smallskip}\hline
\end{tabular}
\end{table}

\begin{figure}
  \includegraphics{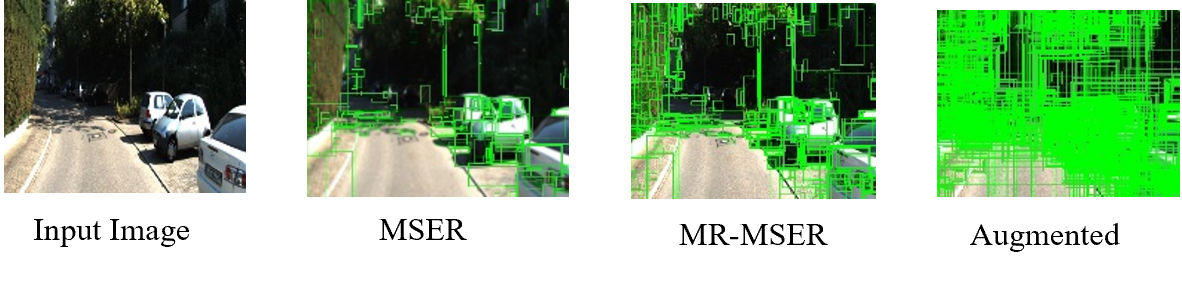}
\caption{Results of applying MSER, MR-MSER, and augmented MR-MSER over KITTI vehicle images}
\label{fig:10}       
\end{figure}

\begin{table}
\caption{EVALUATION OF VEHICLE DETECTION USING THE UA-DETRAC DATASET}
\label{tab:2}       
\begin{tabular}{llll}
\hline\noalign{\smallskip}
Measures & Easy(AP) & Med(AP) & Hard(AP)  \\
\noalign{\smallskip}\hline\noalign{\smallskip}
Method with MSER & 68.48 & 50.22 & 31.87 \\
Method with MR-MSER & 73.69 & 61.15 & 42.12 \\
ACM proposed work & 89.22 & 77.43 & 60.22 \\
\noalign{\smallskip}\hline
\end{tabular}
\end{table}

\begin{figure}
  \includegraphics{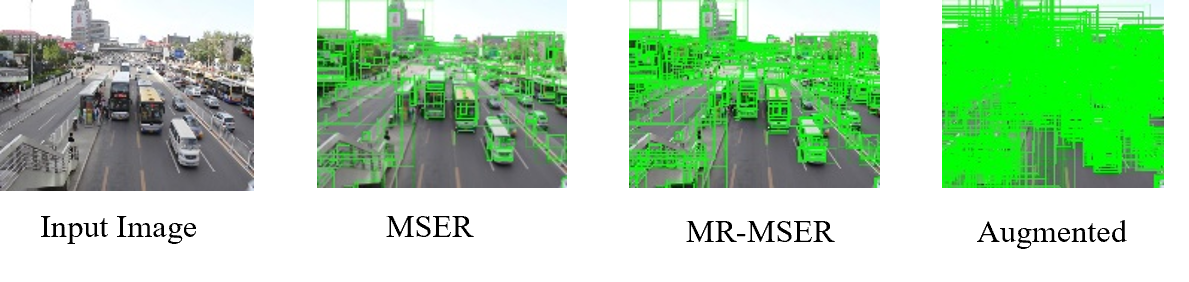}
\caption{Results of applying MSER, MR-MSER, and augmented MR-MSER over UA-DETRAC vehicle images}
\label{fig:11}       
\end{figure}

\begin{table}
\caption{EVALUATION OF VEHICLE DETECTION USING OUR DATASET}
\label{tab:3}       
\begin{tabular}{lll}
\hline\noalign{\smallskip}
Measures & AP & Input image size  \\
\noalign{\smallskip}\hline\noalign{\smallskip}
Method with MSER & 70.87 & 512*512 \\
Method with MR-MSER & 81.65 & 512*512 \\
ACM proposed work & 94.20 & 512*512 \\
\noalign{\smallskip}\hline
\end{tabular}
\end{table}

\begin{figure}
  \includegraphics{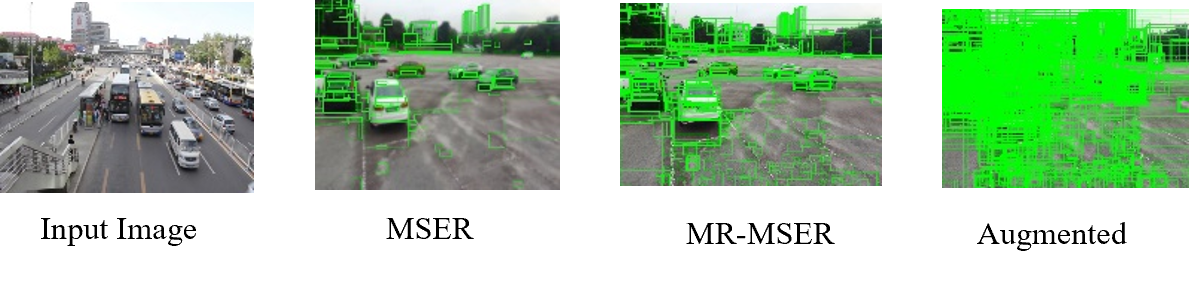}
\caption{Results of applying MSER, MR-MSER, and augmented MR-MSER over our dataset}
\label{fig:12}       
\end{figure}

\subsection{Experiment on vehicle detection}
\label{sec:7}
To make a fair comparison between the proposed ACM method and existing algorithms, average precision (AP0.7) is used for all the experiments in this work [22]. The best match was calculated using ground truth and bounding boxes with a threshold of 0.7 for the correct count (true positive). Otherwise, it is considered a false negative. To evaluate the performance of the proposed method over the KITTI dataset, the number of frames processed per second (FPS) is measured. As the inference speed is related to the hardware equipment, the inference test FPS in this paper is carried out on an NVIDIA RTX 2080Ti. Table IV evaluates both the performance and the efficiency of ACM's proposed method over different methods.

\begin{table}
\caption{EVALUATION OF VEHICLE DETECTION USING THE KITTI DATASET}
\label{tab:4}       
\begin{tabular}{llll}
\hline\noalign{\smallskip}
Measures & AP0.7 & FPS & Image input size  \\
\noalign{\smallskip}\hline\noalign{\smallskip}
MS-CNN\cite{Ref24} & 87.42 & 8.13 & 1920*576 \\
SINet\cite{Ref28} & 89.82 & 23.98 & 1920*576 \\
Faster R-CNN & 59.7 & - & - \\
YoloV3\cite{Ref25} & 79.49 & 43.57 & 512*512 \\
YoloV4\cite{Ref26} & 90.5 & 52.14 & 512*512 \\
ACM Proposed work & 90.08 & 60.42 & 512*512 \\
\noalign{\smallskip}\hline
\end{tabular}
\end{table}

\begin{figure}
  \includegraphics{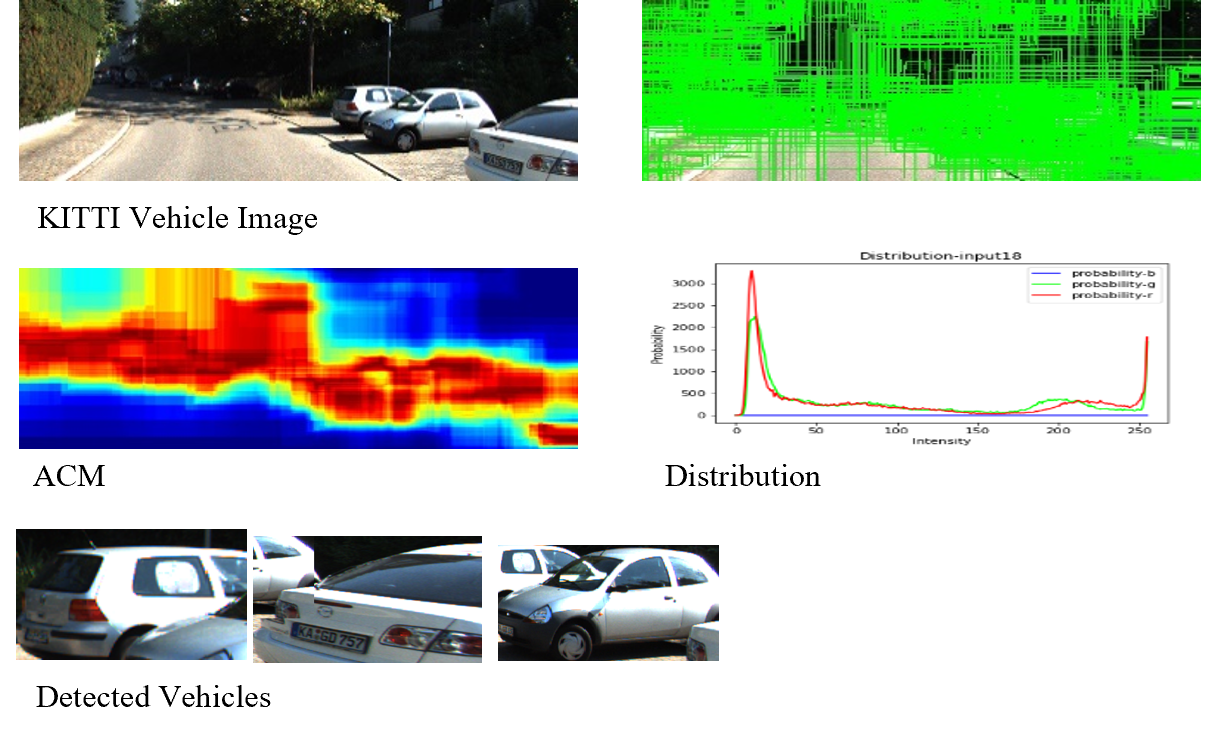}
\caption{Results of applying the proposed ACM to the KITTI sample image.}
\label{fig:13}       
\end{figure}

To evaluate the detection accuracy for different challenge scenarios, the UA-DETRAC dataset is used with different levels of difficulty.

\begin{table}
\caption{EVALUATION OF VEHICLE DETECTION USING THE DETRAC DATASET}
\label{tab:5}       
\begin{tabular}{llll}
\hline\noalign{\smallskip}
Measures & Easy(AP) & Med(AP) & Hard(AP)  \\
\noalign{\smallskip}\hline\noalign{\smallskip}
YoloV3\cite{Ref25} & 83.28 & 62.25 & 42,44 \\
RFCN\cite{Ref27}& 92.32 & 75.67 & 54.31 \\
Yolov5s\cite{Ref29} & 94.2 & - & - \\
Yolov5m\cite{Ref29} & 95.6 & - & - \\
ACM Proposed work & 89.22 & 77.43 & 60.22 \\
\noalign{\smallskip}\hline
\end{tabular}
\end{table}

\begin{figure}
  \includegraphics{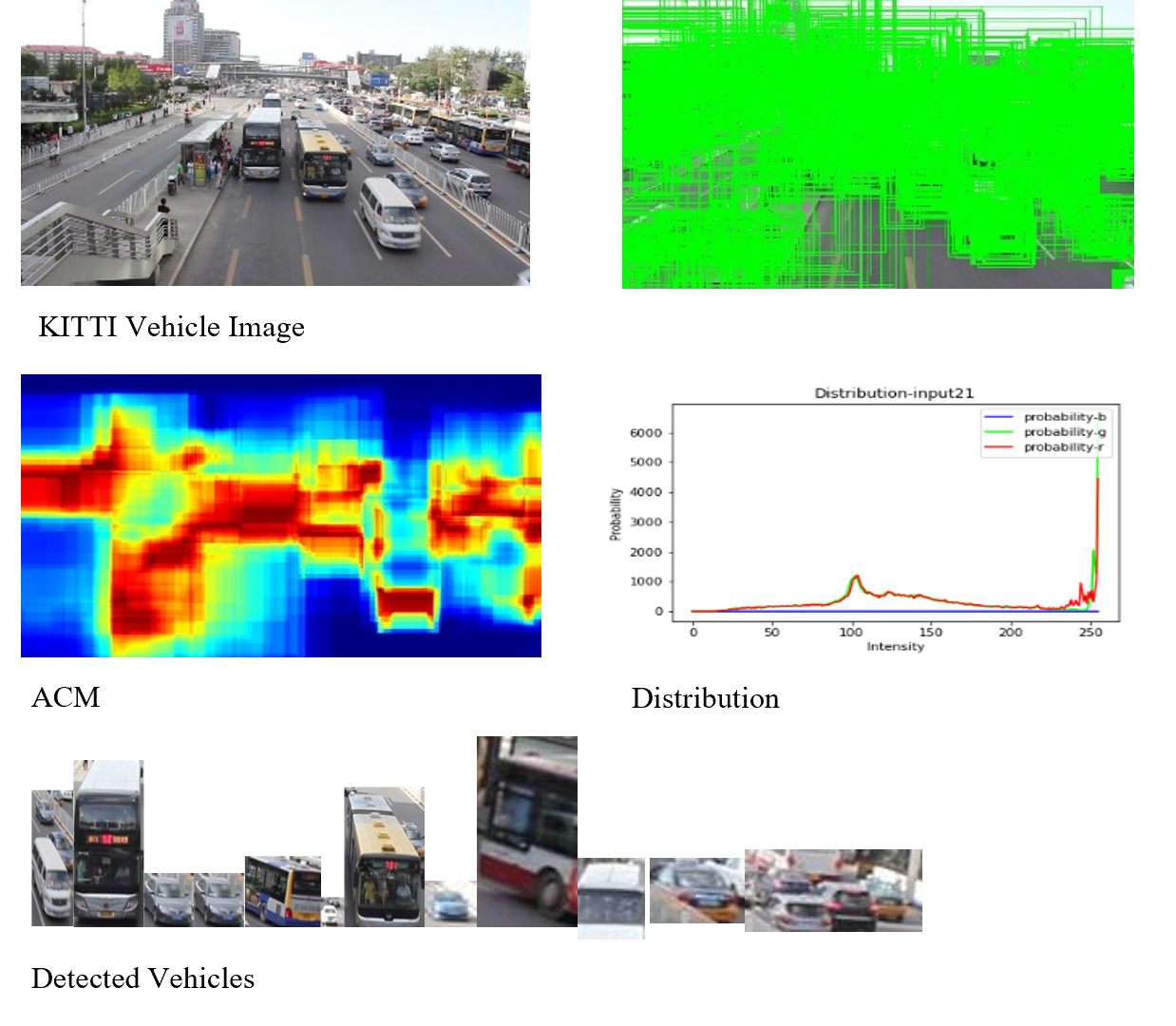}
\caption{Results of applying the proposed ACM to the DETRAC sample image}
\label{fig:14}       
\end{figure}

We also implemented different models of YOLO on the collected dataset to evaluate both the detection performance and the accuracy. The performance is tested using the same hardware used for the KITTI dataset
\begin{table}
\caption{EVALUATION OF VEHICLE DETECTION USING OUR COLLECTED DATASET}
\label{tab:6}       
\begin{tabular}{llll}
\hline\noalign{\smallskip}
Measures & AP0.7 & FPS & Image input size  \\
\noalign{\smallskip}\hline\noalign{\smallskip}
YoloV4\cite{Ref26} & 96.89 & 52.28 & 512*512 \\
YoloV5S\cite{Ref29} & 97.52 & 55.4 & 512*512 \\
YoloV5M\cite{Ref29} & 97.7 & 33.2 & 512*512 \\
ACM Proposed work & 94.2 & 61.20 & 512*512 \\
\noalign{\smallskip}\hline
\end{tabular}
\end{table}

\section{Conclusion}
\label{sec:8}
In this work, we proposed a new method for vehicle detection. We introduced new concepts, namely, the augmented confidence map (ACM), which was fused with traditional learning algorithms to achieve a faster classification method on top of enhanced features generated by a simple CNN. The proposed system shows the capability of detecting vehicles in different scenarios with faster performance. Experimental results conducted on benchmark data called KITTI and DETRAC and on our drone-captured datasets show that the proposed system outperforms the existing methods in terms of timing and can achieve accurate results near YOLOv4. Furthermore, the proposed system is consistent for different datasets and situations.

%
%



\end{document}